\pdfoutput=1
\documentclass[11pt]{article}

\usepackage[preprint]{acl}

\usepackage{times}
\usepackage{latexsym}
\usepackage{adjustbox}

\usepackage[T1]{fontenc}

\usepackage[utf8]{inputenc}

\usepackage{microtype}

\usepackage{inconsolata}

\usepackage{graphicx}

\usepackage[notes=false, done=false, later=true,camera=true]{dtrt}

\def\prefixesnlimodelname{MiniTruePrefixes}
\def\trueteacher{TrueTeacher}
\def\ragtruth{RAGTruth}
\def\summedits{SummEdits}
\def\minitrue{MiniTrue}
\def\minicheck{MiniCheck}
\def\mauve{MAUVE}
\def\true{TRUE}
\def\summeditsprefixes{\summedits Prefixes}
\def\ragtruthprefixes{\ragtruth Prefixes}

\def\shortprefixesnlimodelname{MTP}
\usepackage{amsmath}
\usepackage{booktabs} 

\title{PrefixNLI: Detecting Factual Inconsistencies as Soon as They Arise}

\author{
    Sapir Harary$^{1}$ \quad
    Eran Hirsch$^{1}$ \quad
    Aviv Slobodkin$^{1}$ \quad
    David Wan$^{2}$ \quad
    Mohit Bansal$^{2}$ \quad
    Ido Dagan$^{1}$ \\
    $^{1}$Bar-Ilan University \quad $^{2}$UNC Chapel Hill \\
    \texttt{\{sapirharary1, hirsch.eran, lovodkin93\}@gmail.com} \\
    \texttt{\{davidwan, mbansal\}@cs.unc.edu} \quad \texttt{dagan@cs.biu.ac.il}
}

\definecolor{darkgreen}{rgb}{0.0, 0.5, 0.0}

\usepackage{multirow}

\begin{document}

\maketitle

\begin{abstract}
Natural Language Inference (NLI) models have been used in various ways to improve the factuality of LLM outputs. This is typically done by applying an NLI model to judge whether the model output is entailed from the supposed evidence, triggering some corrective actions, such as beam reranking at inference time or RL rewards during training. While NLI models are trained to detect factual inconsistencies over complete sentences, 
decisions in the common autoregressive generation architecture are made for each evolving text prefix, during decoding. Addressing this setting, 
we generalize the entailment detection task to apply over arbitrary text \textit{prefixes}, and suggest its utility for improving generation faithfulness.
Providing suitable evaluation and training datasets for this task, we train \textit{\prefixesnlimodelname{}}, a novel specialized model that better detects factual inconsistencies over text prefixes, outperforming comparable baseline NLI models by 5-14 F1 points in prefix-level entailment.
We further demonstrate that integrating \prefixesnlimodelname{} into a controlled decoding framework substantially improves factual consistency in abstractive summarization. When guided by \prefixesnlimodelname{}, LLaMA-3.2-3B-Instruct matches the faithfulness and runtime of the 8B model from the same model family, while using only half the memory.\footnote{We release our code, datasets, and trained models at \url{https://github.com/sapirharary/PrefixNLI}.}
\end{abstract}

\section{Introduction}\label{sec:intro}

Large Language Models (LLMs) have made remarkable progress in text generation, yet they remain prone to generating factually inconsistent statements, known as hallucinations \citep{mishra2024finegrained}. In this paper, we focus on generation scenarios where the model output is expected to be factually consistent with supporting textual evidence. Such evidence may be explicitly provided as part of the task input, as in text summarization, or may be retrieved in a Retrieval Augmented Generation (RAG) setting  \citep{lewis2020rag,Jizacard2023atlas}.

Directing LLM generation to be factually consistent with the given inputs is challenging. A notable research line addresses this goal by providing the LLM with factual consistency feedback over the generated text. This is commonly done by employing a Natural Language Inference (NLI) model \citep{dagan2005pascal, bowman-etal-2015-large}, which classifies whether the generated text, considered as the hypothesis, is entailed by the given source texts, considered as the premise. Such entailment feedback has been provided either at training time, by incorporating the entailment score as a Reinforcement Learning (RL) reward \citep{roit-etal-2023-factually}, or at inference time, by utilizing the entailment score for reranking the decoding beam during generation, within a controlled decoding scheme \citep{wan2023faithfulness,sridhar2023improvedbeamsearchhallucination}.
\begin{figure*}[t!]
    \centering
    \includegraphics[width=0.95\textwidth]{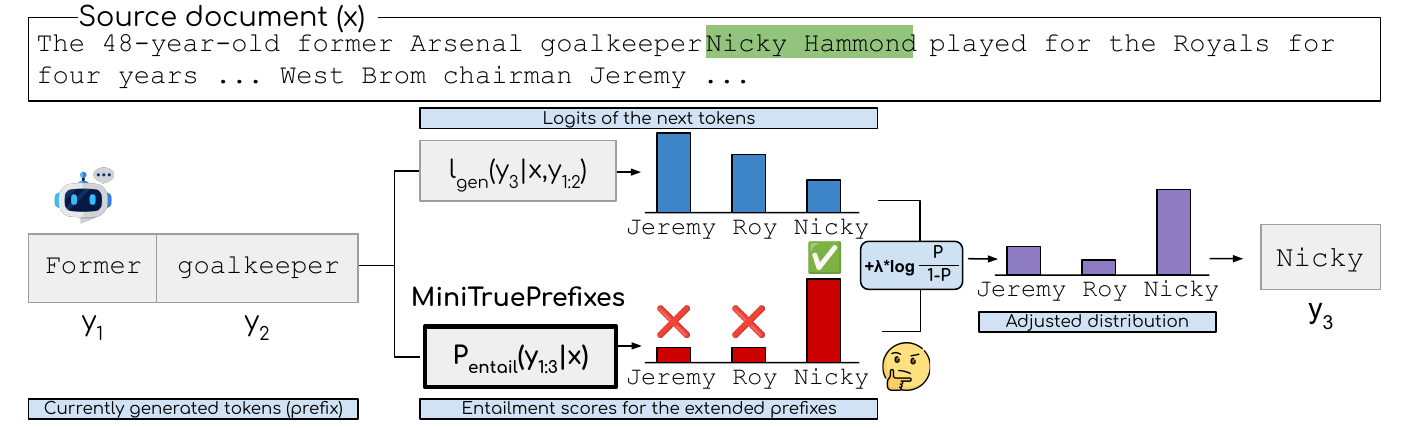}
    \caption{Illustration of PrefixNLI and its downstream utilization for controlled decoding. During autoregressive generation, the base model favors a hallucinated token (“Jeremy”) that is not supported by the source document. Our \prefixesnlimodelname{} model directly evaluates the factual consistency of candidate prefixes at each generation step, assigning here a low entailment score to this unfaithful continuation. For tokens with low entailment probability, we apply a penalty, effectively discouraging unfaithful continuations. This guides generation toward faithful outputs (“Nicky”) in a fine-grained and efficient manner.}
    \label{fig:cd_figure}

\end{figure*}
Since LLM generation is predominantly autoregressive, generating one token at a time, it would be appealing to provide the model with factual consistency feedback at each generation stage, that is, over sentence \textit{prefixes}. However, the entailment recognition task has been originally defined for hypotheses that consist of one or more \textit{complete} sentences, and NLI models have been accordingly trained over datasets with such hypotheses. This discrepancy has led to certain compromises in prior work. In the RL setting, the entailment RL reward was provided only at the end of complete sentences, missing the opportunity for earlier and more granular feedback. In the controlled decoding framework, prior methods greedily completed text prefixes during generation to complete texts (a ``lookahead''), which were only then scored for entailment; this introduced noise in judging entailment for the prefix itself, while also incurring significant computational costs (see~\S\ref{sec:background}).

In this paper, we propose providing LLMs with entailment scoring feedback that is computed \textit{directly} for each text prefix during the autoregressive generation. To that end, we first introduce the \textit{PrefixNLI} task, which extends the traditional textual entailment definition to apply over arbitrary text prefixes as hypotheses, and introduce suitable test and training datasets for this task (\S\ref{sec:prefixes_nli_model}). Next, we train a dedicated NLI model for the PrefixNLI task, \prefixesnlimodelname{} (\S\ref{sec:nli_model}), and show, in an intrinsic evaluation, that it significantly outperforms standard NLI models on prefix-level inference, with relative improvements of over 5 and 14 F1 points across two prefix-level entailment evaluation sets (\S\ref{sec:prefix_results}).
Finally, we apply our \prefixesnlimodelname{} model within a controlled decoding framework, showing consistent faithfulness gains across model sizes and datasets on the abstractive summarization task, including a +5.5 faithfulness points improvement over the LLaMA-3.2-8B-Instruct baseline \citep{grattafiori2024llama3herdmodels}, while remaining much faster than prior controlled decoding methods (\S\ref{sec:cd}). 
For memory-efficient settings, we match the performance and speed of LLaMA-3.2-8B-Instruct with half the memory consumption.
These contributions are illustrated in Figure \ref{fig:cd_figure}. In light of our results, we suggest the broader potential of enhancing faithfulness in text generation via prefix-based NLI, including its incorporation within token-level reinforcement learning in future research.

\begin{figure*}[t]
  \centering
        \includegraphics[width=0.9\textwidth]{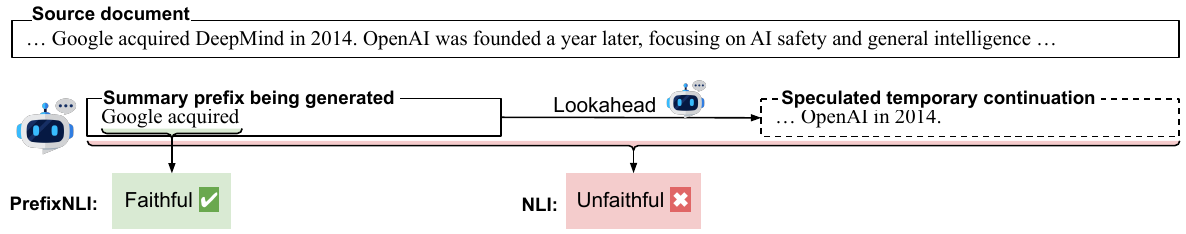}

\caption{Comparing faithfulness assessment approaches in controlled decoding. PrefixNLI evaluates the faithfulness of the generated prefix itself. In contrast, lookahead-based methods \citep{wan2023faithfulness,sridhar2023improvedbeamsearchhallucination} first complete the summary before evaluation, which would be misleading in case the completed summary is found unfaithful due to a factual inconsistency that arises only within the speculated completion, as illustrated in the figure.}
  \label{fig:prefix-nli-hallucination}
\end{figure*}

In summary, our main contributions include: 
\begin{itemize}
    \item Introducing the \textbf{PrefixNLI} task and accompanying datasets, extending natural language inference to arbitrary text prefix hypotheses. 
    \item Developing \prefixesnlimodelname{}, the first entailment model trained specifically for prefix-level inference, establishing a strong baseline on our proposed benchmark. 
    \item Integrating prefix entailment into a \emph{controlled decoding} framework, demonstrating significant factual consistency improvements while maintaining efficiency.
\end{itemize}

\section{Background} \label{sec:background}

A persistent challenge in grounded natural language generation tasks such as summarization \citep{maynez-etal-2020-faithfulness,mckenna-etal-2023-sources} and reference-based question answering \citep{zhang2023languagemodelhallucinationssnowball} is avoiding \textit{factual inconsistencies}, that is, generating text fragments that are not supported by the grounding source (``unfaithful").

Prior work has explored various strategies for improving the faithfulness of text generation, broadly categorized into training-time, generation-time, and post-generation methods. Post-generation methods, such as RARR \citep{gao-etal-2023-rarr}, detect and revise unsupported spans only \textit{after} generation, possibly using retrieved evidence. 

Potentially more appealing approaches aim to direct models to generate faithful outputs up front. Within training-time approaches, this has been done by modifying the model’s learning objective \citep{roit-etal-2023-factually,tian2023fine} or by improving the training data \citep{wan-bansal-2022-factpegasus}. Generation-time methods integrated faithfulness estimation into the decoding process, and utilized it to modify the ranking of candidate paths within the decoding beam, following the general paradigm of controlled decoding \citep{deng-raffel-2023-reward, yang-klein-2021-fudge}, which has recently been applied for improving factual faithfulness \citep{wan2023faithfulness, sridhar2023improvedbeamsearchhallucination}.
An alternative generation-time approach, context-aware decoding \citep{shi-etal-2024-trusting}, encourages the model to rely more on the input context rather than prior knowledge, by contrasting the model’s output probabilities with and without the context input.

A prominent approach for estimating faithfulness, in either training or decoding time, has been to employ NLI models.
\citet{roit-etal-2023-factually} utilized such a model to compute sequence-level faithfulness rewards within a reinforcement learning framework, encouraging faithful generation. Since NLI models are inherently trained to consider hypotheses consisting of complete sentences, the reward was applied only at the end-of-sequence (EOS) token, assigning a zero reward to all other tokens.
For a similar reason, \citet{wan2023faithfulness} and  \citet{sridhar2023improvedbeamsearchhallucination} applied a faithfulness score 
to candidate decoding beam paths only after applying a lookahead mechanism, which greedily generated a (temporary) full summary completion for each candidate token in the beam. Thus, the entailment estimation was applied not only to the currently examined prefix at each decoding step, but rather to a speculated complete summary that augments the current prefix. This method suffers from two substantial limitations. First, the decoding process becomes very costly computationally, since a complete temporary summary has to be generated for each candidate token at each generation step. Second, the entailment estimation is noisy, as it is not applied solely to the currently examined candidate prefix, but rather to the full speculated summary that was generated ad-hoc to complete the examined prefix. Thus, if the full summary is found to be unfaithful, it is not known whether the factual inconsistency exists already in the examined prefix or only in its speculated continuation. In the latter cases, prefixes that are still faithful, and might eventually lead to faithful summaries, get penalized unjustifiably, as illustrated in Figure \ref{fig:prefix-nli-hallucination}. In our work, we aim to circumvent these two deficiencies of the prior lookahead approach by applying a suitable NLI model directly over the beam prefixes.

\section{PrefixNLI: Task and Datasets}\label{sec:prefixes_nli_model}
NLI models are widely used to assess factual consistency in text generation, but their utility in providing feedback \textit{during} generation remains underexplored. Motivated by the need to detect factual errors as soon as they emerge, we introduce the \textbf{PrefixNLI} task, which targets entailment detection over incomplete hypothesis texts. This can be used as a reward signal during training or to steer decoding toward faithful generations at inference.

\subsection{Task definition} \label{subsec:task_definition}
Given a premise text $x$ and an arbitrary text prefix, $y_{1:t}$, considered as the
hypothesis, we would like to predict whether the hypothesis is textually entailed by the premise, or not (a binary entailment setting). 
Since a text prefix might be ungrammatical or even nonsensical, we refine the definition of entailment to fit this setting: a text prefix is entailed by a premise if there may be a sensible completion of the prefix to a complete text that would be entailed by the premise (see Figure \ref{fig:prefix-nli-hallucination} for an illustration). Accordingly, if the prefix already includes some details that are not entailed by the premise then the prefix as a whole is considered not entailed.

\subsection{Constructing evaluation and training datasets} \label{subsec:datasets}
To facilitate research on our novel prefix entailment task, we derive evaluation and training datasets from existing textual entailment and factual consistency data. 
We first describe the general methodology used to generate our datasets, with specific details for each dataset and an illustrating example provided in the subsections below. 

For each dataset instance, we first obtained the source document $x$, regarded as the premise, and a hypothesis text $y$ (typically a summary), which may or may not be faithful to the source. In addition, as detailed below, for non-entailed hypotheses we detected the \textit{first} span, denoted by $s$, that expresses information that is factually inconsistent with the source.
Given this span $s$, we can now deduce that all prefixes ending before the starting position of $s$ are entailed by the premise, thus yielding entailed prefix examples. Conversely, all prefixes ending at the ending position of $s$ or later yield non-entailing prefix instances.\footnote{We do not include prefixes that partially contain $s$, since it is impossible to deduce their entailment label from the available data that we leverage.}
Naturally, faithful original hypotheses yield only entailed prefix examples.

\subsubsection{Evaluation benchmark dataset}
Our evaluation benchmark is derived from two diverse factual consistency datasets. Both datasets contain high-quality \textit{human annotations} which either explicitly mark erroneous spans in non-entailed examples or allow such spans to be inferred with high confidence. Specifically, 
\ragtruth{} \citep{niu2024ragtruthhallucinationcorpusdeveloping} contains machine-generated summaries with human marked hallucination spans, while 
\summedits{} \citep{laban2023llmsfactualreasonersinsights} contains human factual consistency verdicts for summaries that were modified by LLMs. \footnote{Dataset statistics are provided in Appendix~\ref{appendix:eval_datasets}.}

\paragraph{\ragtruthprefixes{}}
\ragtruth{} \citep{niu2024ragtruthhallucinationcorpusdeveloping} is a fine-grained hallucination corpus for assessing faithfulness in retrieval-augmented generations. 
\ragtruth{} contains LLM-generated responses with word-level manual annotations marking factually inconsistent spans, which we can directly consider as the unsupported utterance $s$ in our prefix generation methodology (as described above).

We then consider each sentence in the summary separately and generate from it the prefix instances for our dataset.

Overall, we extract 213K prefix instances and set aside 2K instances for the development set.

\paragraph{\summeditsprefixes{}}
\begin{figure*}[t]
    \centering
        \includegraphics[width=\textwidth]{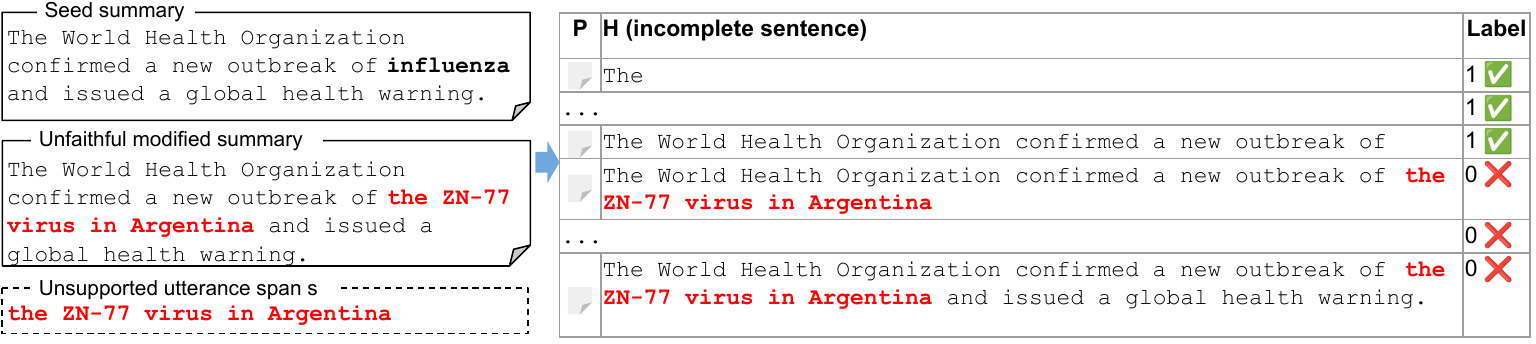}
        \caption{Our \summeditsprefixes{} evaluation dataset creation. Given a seed and a modified summary, we identify the hallucinated span $s$ in an unfaithful summary by identifying where it differs from the seed, highlighted in red. Entailed prefixes are then derived from all positions up to $s$, while non-entailed prefixes are derived starting from the ending position of $s$.}

    \label{fig:summedits_prefixes}

\end{figure*}

Each \summedits{} instance \citep{laban2023llmsfactualreasonersinsights} includes a source document and two summaries: a faithful seed (original) and a modified one generated by an LLM. The modifications are usually limited to local changes, for example an entity swap, a lexical substitute, negation, etc. Each modified summary is evaluated by a human annotator for factual consistency.

To determine the unsupported utterance $s$ in unfaithful summaries, we find the longest common prefix and suffix between the modified and seed summaries and mark the span in-between as $s$. As demonstrated in Figure \ref{fig:summedits_prefixes}, the unsupported utterance $s$ would be the entire span between \textit{``the''} and \textit{``Argentina''}. 
According to our instance generation methodology, all prefixes preceding $s$ are regarded as entailed, while prefixes starting at the last position of $s$ or later are regarded as not entailed.

We construct our dataset by selecting instances from the \textit{News} domain with up to a single LLM modification.
To avoid length bias, we stratify over the prefix position $t$ such that each prefix length has an equal number of factually consistent and inconsistent examples.

\subsubsection{Training dataset}
\label{subsec:datacreation}

To train our PrefixNLI model, we construct two complementary prefix-level NLI datasets, each motivated by a different goal. The first is derived from the \trueteacher{} dataset \citep{gekhman2023trueteacherlearningfactualconsistency}, which contains LLM-generated summaries labeled for factual consistency. While \trueteacher{} provides only summary-level labels, we adapt it to the prefix setting by identifying hallucinated spans within unfaithful summaries using GPT-4 \citep{openai2024gpt4technicalreport} prompting (see Appendix~\ref{appendix:more_details_trainset_entailment}). This dataset captures naturally occurring LLM hallucinations, offering a realistic distribution of factual errors encountered in practice.

Our second training dataset consists of synthetically generated summaries specifically crafted to contain fine-grained and subtle hallucinations. Generating synthetic summaries allows us to control for the types of hallucination in the data and provide a broad coverage of nuanced errors. These summaries were generated using GPT-4 and the complete prompt is provided in Appendix~\ref{appendix:more_details_trainset_entailment}.

\section{PrefixNLI Model}\label{sec:nli_model}

In this section we introduce \textit{\prefixesnlimodelname{}}, trained for the PrefixNLI task, first describing its architecture (\S\ref{subsec:model_architecture_and_inference}) and then its training regime (\S\ref{subsec:training_regime}).

\subsection{Model architecture and inference} \label{subsec:model_architecture_and_inference}

The primary intended usage of PrefixNLI is to evaluate the consistency of a prefix with the source as it evolves by appending one token at a time, as illustrated in Figure~\ref{fig:cd_figure}.
Such an approach can be expensive as each decoding step requires another entailment inference call per candidate token, e.g. ``Jeremy'', ``Roy'', and ``Nicky'' in the figure.
However, since the model is only generating the last token, the same prefix will be used in subsequent decoding steps, e.g. ``Former goalkeeper''. This step-by-step formulation naturally aligns with decoder-only architectures \citep{grattafiori2024llama3herdmodels}, as used by the entailment model, which are well-suited to this task due to their efficient support for prefix caching. That is, they store and reuse key-value (KV) pairs, which is critical for reducing computational overhead \citep{kwon2023efficient}.

Accordingly, we use the LLaMA-3.2-Instruct model \citep{grattafiori2024llama3herdmodels} as our base model. 
We selected a small and relatively nimble model with 1B parameters in order to further minimize computational costs.
We then follow the entailment classification architecture of the TrueTeacher NLI system \citep{gekhman2023trueteacherlearningfactualconsistency}.

Given a source document $x$ and a hypothesis prefix $y_{1: t}$, the model predicts the entailment decision.
To make a prediction, we construct a prompt from $x$ and $y_{1:t}$ (see Appendix~\ref{appendix:more_train_details} for technical details), and infer entailment if 
the probability for the entailment class (technically for the class label token ``1''), denoted $P_{\text{entail}}(y_{1:t}\mid x)$, is higher than 0.5.

\subsection{Training regime} \label{subsec:training_regime}

We train our model in two stages: we train a summary-level NLI model, \textit{\minitrue{}}, which we then adapt for prefix-level supervision, \textit{\prefixesnlimodelname{}}.
In both stages we use cross-entropy loss between the model’s predicted and target labels. Akin to the training regime of TrueTeacher \citep{gekhman2023trueteacherlearningfactualconsistency}, we provide as input a pair of a premise text and a (truncated) hypothesis text, and train the model to generate the token ``1'' if it is entailed or ``0'' for non-entailed verdicts.

This formulation enables scoring of both truncated and complete hypotheses with the same setup.

\paragraph{Fine-tuning a lightweight entailment model.} 
We fine-tune the base model on the \trueteacher{} dataset \citep{gekhman2023trueteacherlearningfactualconsistency} and the ANLI dataset \citep{nie2020adversarialnlinewbenchmark}, mirroring the training setup of the original \trueteacher{} model with a more parameter efficient variant.
The resulting model, which we refer to as \minitrue{}, achieves strong entailment performance while remaining lightweight and computationally efficient.
See Appendix~\ref{appendix:minitrue_true} for the evaluation of \minitrue{} on the \true{} benchmark.

\paragraph{Adapting for prefix entailment.} We introduce \textit{\prefixesnlimodelname{}}, obtained via an additional fine-tuning step on our curated prefix-level entailment dataset. 

This additional training step enables the model to better handle the linguistic incompleteness and increased ambiguity present in truncated hypotheses, resulting in more reliable factual inconsistencies detection on truncated hypotheses.
See Appendix~\ref{appendix:more_train_details} for more details.

\section{PrefixNLI Intrinsic Evaluation}\label{sec:prefix_results}

\subsection{Experimental setup}

\paragraph{Baselines} We evaluate two baseline models trained for the NLI task. The first is \minitrue{}, our lightweight entailment model described in  \S\ref{subsec:training_regime}, which was trained on standard NLI data and served as the base model for further prefix-level finetuning of the \prefixesnlimodelname{} model. 

The second is \minicheck{} \citep{tang2024minicheckefficientfactcheckingllms}, the current state-of-the-art factual consistency model of similar size (770M parameters). 

\paragraph{Evaluation metrics}
Given an evaluation benchmark consisting of premise $x$, prefix hypothesis $y_{1:t}$, and a gold entailment decision, we compute for each system the micro-averaged F1 score for the \textit{unfaithful class} across all instances.\footnote{We also report the micro-averaged F1 score for the faithful class in Appendix ~\ref{appendix: mtp_additional_metrics}.} This choice aligns with our objective of detecting factual \textit{inconsistencies} as soon as they arise during decoding, hence adopting a metric that directly reflects the model’s ability to identify unfaithfulness. These F1 scores are reported with their 95\% confidence interval.

\subsection{Results} \label{tab:nli_results}

\begin{table}[t]
    \centering
    \small
    \adjustbox{max width=\columnwidth}{

    \begin{tabular}{lcc}
        \toprule
        \textbf{Model} & \textbf{\summeditsprefixes{}} & \textbf{\ragtruthprefixes{}} \\
        \midrule
        \minicheck{} (Flan-T5) & 72.9 $\pm$ 1.0 & 33.3 $\pm$ 0.5 \\
        \minitrue{} & 69.8 $\pm$ 1.2 & 27.5 $\pm$ 0.5 \\
        \prefixesnlimodelname{} & \textbf{78.1} $\pm$ 1.0 & \textbf{47.6} $\pm$ 0.6 \\
        \bottomrule
    \end{tabular}
    }
    \caption{F1 scores for the unfaithful class on \summeditsprefixes{} and \ragtruthprefixes{}.}
    \label{tab:model_comparison}
\end{table}

As shown in Table~\ref{tab:model_comparison}, \prefixesnlimodelname{} outperforms the strongest baseline of similar size, \minicheck{} (Flan-T5), by 5.2 and 14.3 points on \summeditsprefixes{} and \ragtruthprefixes{}, respectively. It also surpasses \minitrue{}, demonstrating the benefit of prefix-level fine-tuning.

These results highlight the benefit of training a targeted prefix-level NLI model, over prefix-level training data, when aiming to detect unfaithful text prefixes. Appendix~\ref{appendix:qualitative_examples} presents a qualitative example where \prefixesnlimodelname{} succeeds while the sentence-level models fail. Additionally, we performed a manual error analysis described in Appendix~\ref{appendix:error_analysis}.

\subsection{Performance by prefix length}
\label{sec:prefix_analysis}

\begin{figure}[t]
    \centering
    \includegraphics[width=\columnwidth]{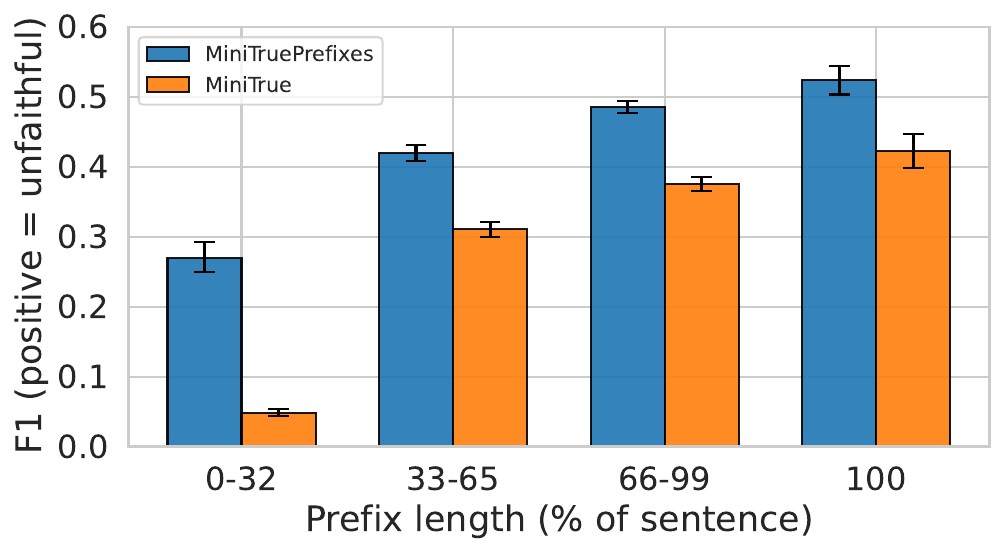}
        \caption{
        F1 scores across prefix-length bins for 
        \prefixesnlimodelname{} and \minitrue{}, 
        with 95\% confidence intervals.
    }
    \label{fig:prefix-f1-bars}
\end{figure}

We analyze how performance is influenced by the length of the prefix relative to the length of the sentence it was extracted from.
Confidence intervals (95\%) were computed per bin using 1{,}000 bootstrap resamples.
The results in Figure~\ref{fig:prefix-f1-bars} show that \prefixesnlimodelname{} consistently outperforms the \minitrue{} baseline, not only on partial prefixes but also on full sentences (100\% bin). The performance gap is particularly pronounced in the earliest generation stage (0-32\% bin), where \prefixesnlimodelname{} achieves a 27.0 F1-score, over 5.5$\times$ higher than the 4.9 achieved by \minitrue{}.

While performance naturally increases with longer prefixes due to richer context, the relative advantage of \prefixesnlimodelname{} remains substantial across all bins, underscoring its stability and robustness.
These results indicate that the model generalizes well across prefix lengths, maintaining reliable detection from early to late generation stages.

\section{Controlled Decoding with PrefixNLI}
\label{sec:cd}

In this section, we show how PrefixNLI models can be used to generate more faithful summaries, with a reasonable computational cost. As motivated in \S\ref{sec:intro} and \S\ref{sec:background}, our approach avoids altogether using the inefficient and potentially noisy lookahead mechanism employed in prior work \citep{wan2023faithfulness,sridhar2023improvedbeamsearchhallucination}. 

\subsection{Method} \label{subsec:cd_method}

Standard autoregressive decoding methods, including beam search, prioritize \textbf{likelihood}, which often indirectly reflects faithfulness but does not explicitly enforce it.

To improve faithfulness, our method integrates entailment probability into next-token decoding decisions, following the above-mentioned line of recent work. Specifically, we use our \prefixesnlimodelname{} model to calculate the entailment score between the source and the current prefix extended with each candidate next token (as illustrated in Figure \ref{fig:cd_figure}). 
To steer generation away from hallucinations, we penalize tokens with low entailment scores, discouraging unfaithful continuations. This biases the model toward more faithful outputs without fully overriding its original preferences, allowing us to reduce factual inconsistencies as soon as they arise.

Formally, let $\ell_i = \ell_{\text{gen}}(y_t^i \mid x, y_{1:t-1})$ denote the logit assigned by the language model to the $i$-th token in the vocabulary at time step $t$, given the source and current prefix, and let $p_i = p_{\text{entail}}(y_{1:t}^i \mid x)$ denote the entailment score of the resulting extended prefix.

For each top-$p$ candidate token with an entailment score below our tuned rectification threshold $\tau=0.5$ (see Appendix~\ref{appendix:rectification_threshold}), indicating that non-entailment is more likely than entailment, we update its logit as follows:
\[
\ell_i \leftarrow \ell_i + \lambda \cdot \log\left(\frac{p_i}{1 - p_i}\right)
\]

The adjustment term is defined as the scaled log-odds of the entailment probability. It serves as a penalty, taking negative values when $p_i < 0.5$, with increasing magnitude as $p_i$ decreases. $\lambda$ is a scaling factor that controls the strength of the penalty. Tokens with $p_i > 0.5$ remain unchanged, thereby preserving the original distribution while discouraging unfaithful continuations. All tokens not included in the top-$p$ set are assigned a logit of $-\infty$ and thus excluded from consideration. For beam search with beam size $K$, we maintain the top $K$ candidate sequences based on their cumulative adjusted log-probabilities. Hyper-parameters were tuned on the development set of the CNN/DM, see Appendix~\ref{appendix:cd_hyperparameters} for more details.

\subsection{Experimental setup} \label{subsec:cd_experimental_setup}

\paragraph{Datasets} We follow prior work in this area \citep{wan2023faithfulness, roit-etal-2023-factually, sridhar2023improvedbeamsearchhallucination} and focus our evaluation on abstractive summarization, a typical representative of source-grounded generation settings. We conduct experiments on the XSum \citep{narayan2018dontdetailsjustsummary} and CNN/DM \citep{hermann2015teachingmachinesreadcomprehend} datasets, using 2,500 test set documents from each.

\paragraph{Baselines} Our experimental setup includes two components: the generator LLM, used to generate the outputs, and the NLI model which is employed in the decoding process to estimate the faithfulness of candidate tokens. For the generator LLM, we conduct experiments with several models from the LLaMA-3 series: LLaMA-3.2-1B-Instruct, LLaMA-3.2-3B-Instruct, and LLaMA-3.2-8B-Instruct \citep{grattafiori2024llama3herdmodels}, in a zero-shot setting (see Appendix~\ref{appendix:generation_prompt_cd}).

For the decoding process intervention, we compare the following approaches: \textbf{Vanilla} denotes standard generation without intervention in the decoding process. \textbf{Lookahead} implements the lookahead algorithm of \citet{wan2023faithfulness}. We greedily extend the current beam prefix to a complete (temporary) summary (as described in \S\ref{sec:background}), using the same model as the LLM generator model, and then evaluate the faithfulness of this full summary using our \minitrue{} entailment model. \textbf{CAD} compares our method to the Context-Aware Decoding (CAD) algorithm proposed by \citet{shi-etal-2024-trusting}, which does not rely on an NLI metric (additional setup details are provided in Appendix~\ref{appendix:cad_details}). \textbf{Prefix} applies our proposed controlled decoding method (\S\ref{subsec:cd_method}), while utilizing our \prefixesnlimodelname{} as the prefix entailment model.

\subsubsection{Metrics} \label{subsec:evaluation}
We evaluate the complete generated summaries using the following automatic metrics:
\paragraph{Faithfulness} Following recent works \citep{wan2025mammrefinerecipeimprovingfaithfulness,lee2024unisumevalunifiedfinegrainedmultidimensional}, we evaluate faithfulness using the \minicheck{} entailment model \citep{tang2024minicheckefficientfactcheckingllms}. We use Bespoke-MiniCheck-7B, the current state-of-the-art for this task. Each summary sentence is individually compared against the entire source document and assigned a binary entailment label, following \cite{tang2024minicheckefficientfactcheckingllms}, binary labels were used for evaluation. The proportion of entailed sentences in each summary is regarded as its faithfulness score, then averaged over all evaluated summaries.

We also evaluate faithfulness using GPT-4.1\footnote{Evaluation based on 1,000 documents per dataset.} \citep{openai2024gpt4technicalreport} with a prompting setup adapted from \citet{wadhwa-etal-2024-learning-refine} (described in Appendix~\ref{appendix:gpt_faithfulness}). Both this method and \minicheck{} have been shown to correlate strongly with human judgments of faithfulness \citep{tang2024minicheckefficientfactcheckingllms, liu-etal-2023-g, chiang-lee-2023-large, gao2023humanlikesummarizationevaluationchatgpt}.

\paragraph{Content Quality} We assess output quality using standard summarization metrics. Specifically, we report ROUGE-L F1 \citep{lin-2004-rouge} against reference summaries and evaluate fluency with \mauve{} \citep{pillutla2021mauvemeasuringgapneural}, using a scaling factor of $c=0.5$, to ensure our decoding does not harm fluency.
\paragraph{Latency} Average generation speed (in seconds) per summary, measuring the cost of applying the NLI model at each step. Evaluated on a single NVIDIA A100 80GB GPU.\footnote{For more details, see Appendix~\ref{appendix:exp_setup_cd}.} For completeness, Appendix~\ref{appendix:comp_cost_analysis} reports the theoretical floating-point operations (FLOPs) per-token cost associated with our method.

We report the standard error of the mean when applicable to assess statistical significance.

\subsection{Results}\label{subsec:cd_results}
\begin{table*}[!t]
    \centering
    \small
    \resizebox{\textwidth}{!}{
    \begin{tabular}{l | cc | ccc || cc | ccc}
        \toprule
        \multirow{3}{*}{Model} 
        & \multicolumn{5}{c||}{\textbf{XSum}} 
        & \multicolumn{5}{c}{\textbf{CNN/DM}} \\
        \cmidrule{2-11}
        & \multicolumn{2}{c|}{Faithfulness} & \multicolumn{3}{c||}{Auxiliary Metrics} 
        & \multicolumn{2}{c|}{Faithfulness} & \multicolumn{3}{c}{Auxiliary Metrics} \\
        & \minicheck{}$\uparrow$ & GPT-4$\uparrow$ & R-L$\uparrow$ & \mauve{}$\uparrow$ & Speed$\downarrow$
        & \minicheck{}$\uparrow$ & GPT-4$\uparrow$ & R-L$\uparrow$ & \mauve{}$\uparrow$ & Speed$\downarrow$ \\
        \midrule
        Vanilla (1B) 
            & 66.7 $\pm$ 0.6 & 2.53 $\pm$ 0.02 & 14.4 $\pm$ 0.1 & 91.4 & \textbf{1.74} 
            & 71.7 $\pm$ 0.6 & 3.23 $\pm$ 0.03 & 18.9 $\pm$ 0.1 & \textbf{79.6} & \textbf{1.93} \\
        Vanilla (3B) 
            & 76.2 $\pm$ 0.6 & 3.34 $\pm$ 0.03 & 14.3 $\pm$ 0.1 & \textbf{91.6} & 3.10 
            & 80.1 $\pm$ 0.6 & 3.68 $\pm$ 0.03 & 18.9 $\pm$ 0.1 & 79.3 & 4.46 \\
        Vanilla (8B) 
            & 81.5 $\pm$ 0.5 & 3.63 $\pm$ 0.03 & \textbf{14.9} $\pm$ 0.1 & 91.3 & 5.30 
            & 87.9 $\pm$ 0.6 & 3.87 $\pm$ 0.02 & 19.0 $\pm$ 0.1 & 79.4 & 6.76 \\
        \midrule
        Lookahead (1B) 
            & 70.9 $\pm$ 0.6 & 3.03 $\pm$ 0.03 & 12.8 $\pm$ 0.1 & 89.4 & 114.1 
            & 75.9 $\pm$ 0.6 & 3.28 $\pm$ 0.03 & 17.9 $\pm$ 0.1 & 79.1 & 158.9 \\
        CAD (1B) 
            & 70.1 $\pm$ 0.6 & 3.05 $\pm$ 0.03 & 12.4 $\pm$ 0.1 & 90.8 & 13.3 
            & 75.5 $\pm$ 0.5 & 3.34 $\pm$ 0.02 & \textbf{20.4} $\pm$ 0.1 & 79.4 & 20.0 \\
        \midrule
        Prefix (1B, \shortprefixesnlimodelname{}) 
            & 74.8 $\pm$ 0.6 & 3.13 $\pm$ 0.03 & 13.6 $\pm$ 0.1 & 89.6 & 4.84 
            & 79.2 $\pm$ 0.6 & 3.41 $\pm$ 0.02 & 19.8 $\pm$ 0.1 & 79.1 & 5.67 \\
        Prefix (3B, \shortprefixesnlimodelname{}) 
            & 82.4 $\pm$ 0.5 & 3.59 $\pm$ 0.03 & 14.1 $\pm$ 0.1 & 91.2 & 5.43 
            & 85.8 $\pm$ 0.4 & 3.73 $\pm$ 0.02 & 19.0 $\pm$ 0.1 & 79.1 & 7.18 \\
        Prefix (8B, \shortprefixesnlimodelname{}) 
            & \textbf{87.0} $\pm$ 0.4 & \textbf{3.84} $\pm$ 0.03 & 14.5 $\pm$ 0.1 & 91.3 & 7.23 
            & \textbf{90.8} $\pm$ 0.3 & \textbf{3.90} $\pm$ 0.02 & 18.9 $\pm$ 0.1 & 79.1 & 9.20 \\
        \bottomrule
    \end{tabular}
    }
    \caption{
    Performance across the XSum and CNN/DM datasets for \textbf{LLaMA} models. \textbf{\prefixesnlimodelname{}} is abbreviated as \textbf{\shortprefixesnlimodelname{}}.
    }
    \label{tab:cd_results}
\end{table*}
The results are shown in Table~\ref{tab:cd_results} and our main takeaways are discussed below.

\paragraph{\prefixesnlimodelname{} improves faithfulness across different-sized generators.}
Using our \prefixesnlimodelname{} with the 1B generator yields 7.5-point and 8-point faithfulness improvements (\minicheck{} column) in the CNN/DM and XSum datasets, respectively.

Scaling the generator consistently improves faithfulness: combining our 1B \prefixesnlimodelname{} with a 3B generator improves faithfulness by 5.7 points on CNN/DM and 6.2 points on XSum over the vanilla 3B model. Remarkably, on XSum, our 3B+1B configuration outperforms the vanilla 8B model by 0.9 \minicheck{} points and offers a substantially more compute-efficient alternative, particularly in resource-constrained settings such as on-device deployment. Similarly, using our method with an 8B generator further improves faithfulness by 2.9 points on CNN/DM and 5.5 points on XSum compared to vanilla 8B. These results indicate that prefix-level entailment reranking provides consistent and complementary gains even when applied to strong generator models, and can effectively close the faithfulness gap between model scales. We observe consistent gains in GPT-4 assessed faithfulness across all model sizes using our method.
Examples illustrating how our method fixes or avoids hallucinations in the vanilla summary are presented in Table~\ref{tab:example_predictions} in Appendix~\ref{appendix:improvement_examples}. 

\paragraph{Using \prefixesnlimodelname{} incurs a reasonable computational cost.} The generation time increases by up to 2.9$\times$ for the 1B generator. However, the relative overhead decreases with model size, dropping to 1.4$\times$ for the 8B model, as the fixed-cost entailment model accounts for a smaller proportion of total compute.

\paragraph{\prefixesnlimodelname{} outperforms baseline approaches.}
Both the prior lookahead method and CAD improve faithfulness over vanilla decoding but underperform noticeably compared to our approach. For lookahead, we conjecture that scoring entire speculative continuations, rather than the prefix generated up to the current step, introduces noise into prefix-level entailment judgments and limits effectiveness (see Figure~\ref{fig:prefix-nli-hallucination}). As expected, the lookahead approach is substantially more computationally expensive, running on average 25.8$\times$\ slower than our method. The CAD baseline, which does not rely on an NLI signal, offers meaningful improvements but remains less effective than our entailment-guided approach. It also incurs higher computational costs than ours, with 2.7$\times$ slower runtime on XSum and 3.5$\times$ on CNN/DM. As an ablation, we also evaluated the use of \minitrue{} as the prefix entailment model and found that it consistently underperformed relative to \prefixesnlimodelname{} across all evaluation metrics and model sizes. Complete ablation results are reported in Appendix~\ref{appendix:ablation_study}.

\paragraph{Summary quality is not compromised.} With respect to summary quality, on the CNN/DM dataset our more faithful generation method for the 1B generator yields more relevant summaries with respect to the ROUGE score (+0.9 points). This makes sense since reference summaries in CNN/DM are generally faithful to the source document, hence hallucinations in the predicted summary also reduce similarity to the reference. ROUGE scores for 3B and 8B models show only minor nonsignificant changes when applying our method. On the XSum dataset, the ROUGE relevance score slightly decreases, which is sensible, since XSum references are known to include many factual inconsistencies relative to the source document (by construction, XSum references are not actual summaries) \citep{maynez-etal-2020-faithfulness}. Fluency (\mauve{}) is maintained across models, with negligible differences relative to the vanilla baseline.

\subsection{Results using OLMo generators}
To evaluate the robustness of our method across model families and ensure that the computational overhead remains reasonable even when the generator and entailment model employ different tokenizers, we conducted the \textbf{Prefix} experiment using OLMo models \citep{groeneveld-etal-2024-olmo} of sizes 1B and 7B as generators.

\begin{table*}[!htbp]
    \centering
    \small
    \resizebox{\textwidth}{!}{
    \begin{tabular}{l | cc | ccc || cc | ccc} 
        \toprule
        \multirow{3}{*}{Method} 
        & \multicolumn{5}{c||}{\textbf{XSum}} 
        & \multicolumn{5}{c}{\textbf{CNN/DM}} \\
        \cmidrule{2-11}
        & \multicolumn{2}{c|}{Faithfulness} & \multicolumn{3}{c||}{Auxiliary Metrics} 
        & \multicolumn{2}{c|}{Faithfulness} & \multicolumn{3}{c}{Auxiliary Metrics} \\
        & \minicheck{}$\uparrow$ & GPT-4$\uparrow$ & R-L$\uparrow$ & \mauve{}$\uparrow$ & Speed$\downarrow$
        & \minicheck{}$\uparrow$ & GPT-4$\uparrow$ & R-L$\uparrow$ & \mauve{}$\uparrow$ & Speed$\downarrow$ \\
        \midrule
        Vanilla (1B) 
            & 68.3 $\pm$ 0.6 & 2.92 $\pm$ 0.03 & \textbf{13.1} $\pm$ 0.1 & \textbf{92.3} & \textbf{2.65} 
            & 72.2 $\pm$ 0.5 & 3.38 $\pm$ 0.03 & 17.2 $\pm$ 0.1 & 79.4 & \textbf{3.10} \\
        Vanilla (7B) 
            & 85.2 $\pm$ 0.4 & 3.60 $\pm$ 0.03 & 11.4 $\pm$ 0.1 & 90.2 & 7.43 
            & 87.9 $\pm$ 0.4 & 3.95 $\pm$ 0.02 & 16.3 $\pm$ 0.1 & \textbf{79.5} & 9.09 \\
        \cmidrule{1-11}
        Prefix (1B, \shortprefixesnlimodelname{}) 
            & 76.1 $\pm$ 0.6 & 3.22 $\pm$ 0.03 & 12.9 $\pm$ 0.1 & 91.8 & 6.62 
            & 78.7 $\pm$ 0.5 & 3.55 $\pm$ 0.02 & \textbf{17.3} $\pm$ 0.1 & 79.4 & 8.31 \\
        Prefix (7B, \shortprefixesnlimodelname{}) 
            & \textbf{87.5} $\pm$ 0.4 & \textbf{3.77} $\pm$ 0.03 & 11.3 $\pm$ 0.1 & 89.8 & 10.63 
            & \textbf{90.3} $\pm$ 0.3 & \textbf{4.02} $\pm$ 0.02 & 16.5 $\pm$ 0.1 & 79.4 & 13.0 \\
        \bottomrule
    \end{tabular}
    }
    \caption{
        Performance across the XSum and CNN/DM datasets for \textbf{OLMo} models. 
        \textbf{\prefixesnlimodelname{}} is abbreviated as \textbf{\shortprefixesnlimodelname{}}.
    }
    \label{tab:cd_results_olmo}
\end{table*}

Results for the \textbf{OLMo} family (Table~\ref{tab:cd_results_olmo}) confirm that prefix-level entailment guidance generalizes beyond LLaMA. Applying \prefixesnlimodelname{} to the 1B OLMo generator improves faithfulness by 7.8 and 6.5 \minicheck{} points on XSum and CNN/DM, respectively, while the 7B model gains an additional 2.3 and 2.4 points over its vanilla counterpart. Our prefix-level entailment guidance is architecture-agnostic and scales smoothly with model capacity. Summary relevance and fluency remain stable across both datasets, indicating that the faithfulness gains do not come at the expense of quality. Runtime overheads remain moderate (2.6$\times$ for 1B, 1.4$\times$ for 7B), showing that the approach is computationally efficient and generalizes well across model families.

Integrating \prefixesnlimodelname{} into decoding consistently improves faithfulness without compromising summary quality or fluency. The gains are robust across model scales, architectures, and datasets, establishing prefix-level entailment guidance as an efficient and general approach to reducing hallucinations and motivating future work on refining prefix-entailment modeling and its integration into autoregressive generation.

\section{Conclusions and Future Work} \label{sec:conclusiond}

We introduced PrefixNLI, the task of assessing factual consistency at the text prefix level, with the primary motivation of detecting factual inconsistencies as soon as they arise during autoregressive text generation. 
Promoting research on this task, we introduced suitable datasets and a targeted model, which showed positive results in both an intrinsic evaluation and in improving generation faithfulness in a controlled decoding framework. Our model also yields dramatic efficiency gains compared to prior lookahead-based approaches, facilitating inference-time factuality control. 

Our work opens up several directions for future research on improving and leveraging prefix-level entailment models. 
First, there is room for improving the core PrefixNLI model, through the creation of richer training data, manually or automatically, as well as enhanced modeling, e.g. by identifying semantic unit boundaries as more reliable ``breakpoints'' for assessing prefix entailment. 
Second, there is potential for improving the faithfulness and efficiency of the controlled decoding method, e.g. by more selective application of PrefixNLI models at reliable time steps, or smarter beam management. 
Third, it is very appealing to incorporate PrefixNLI also into the training regime, possibly extending it from using only sentence-level rewards \citep{roit-etal-2023-factually} to more informative and precise token-level rewards. Finally, prefix-level NLI can benefit additional generation tasks, beyond summarization.

\section*{Ethical considerations}
This work aims to improve the faithfulness of language model outputs, which we view as a desirable objective with positive downstream impact. However, our method depends on predictions from \prefixesnlimodelname{}, an entailment model that can occasionally misclassify unfaithful content as faithful. As a result, models using our approach might produce outputs that appear trustworthy despite containing factual errors, which can mislead users to assume a stronger connection between the source and generated text than what actually exists.

We used AI tools to assist with language refinement. All content was subsequently reviewed and validated by the authors for accuracy and correctness.

\section*{Limitations}
\paragraph{Access to logits.}
Our controlled decoding method relies on access to the output logits of the language model in order to modify the token distribution during inference. However, this requirement poses a constraint on the applicability of our method to closed-source, API-based models that do not expose internal logits. As a result, our approach is currently limited to open or locally accessible models.

\paragraph{Inference overhead.}
The method introduces a moderate increase in inference time, stemming from the entailment computations performed at each generation step. While our experiments demonstrate that this additional cost is justified by the resulting gains in factual consistency, future work may explore strategies to reduce computational overhead. For instance, the entailment model could be applied more selectively, only at key decision points such as punctuation marks or in cases of high model uncertainty.

\paragraph{Language and domain scope.}
Although the LLaMA-3.2-Instruct 1B model used as the backbone of the \prefixesnlimodelname{} model supports multilingual generation, our model was trained exclusively on English data. This choice arises from the availability of high-quality datasets with fine-grained hallucination localization, which are largely limited to English and critical for the evaluation of our method. Likewise, our experiments focus on summarization in the news domain, which, although representative of source-grounded generation, may not capture the full diversity of hallucination phenomena across other tasks or domains. Extending the approach to multilingual and cross-domain settings remains an important direction for future work.

\section*{Acknowledgments}
We thank Paul Roit for valuable comments and insightful discussions that contributed to this work. 
We also thank the reviewers for their constructive feedback and suggestions.
This work was supported by the Israel Science Foundation (grant no. 2827/21), NSF-CAREER Award 1846185, NSF-AI Engage Institute DRL-2112635, and a Google PhD Fellowship. The views contained in this article are those of the authors and not of the funding agency.

\bibliography{references}

\appendix

\begin{table*}[htbp]
    \centering
    \small
    \adjustbox{max width=\textwidth}{
    \begin{tabular}{lccccc}
        \toprule
         & \multicolumn{2}{c}{\# Prefixes} & \# Unique documents & Avg. hallucinatory span length & Avg. Prefix length \\
         \cmidrule(lr){2-3}
         & Faithful & Unfaithful &  &  &  \\
         \midrule
         RAGTruthPrefixes & 194,283 & 16,395 & 880 & 11.8 & 14.7 \\
         SummEditsPrefixes & 4,455 & 4,455 & 25 & 4.2 & 27.6 \\
         \bottomrule
    \end{tabular}
    }
    \caption{We report the
number of prefixes in each dataset (faithful vs. unfaithful), along with the number of unique source documents, the average hallucinatory span length and the average prefix length (in tokens).}
    \label{tab:more_data_stats}
\end{table*}

\section{Training Data Construction Details}
\label{appendix:more_details_trainset_entailment}
This section provides additional details about the construction of our training dataset, expanding upon \S\ref{subsec:datacreation}.
\paragraph{Hallucination detection using GPT-4}
We begin by extracting summaries labeled as unfaithful from the \trueteacher{} dataset \citep{gekhman2023trueteacherlearningfactualconsistency}. For each summary, we prompt GPT-4 to identify and highlight the specific span corresponding to the hallucinated content. Table~\ref{tab:hallucination_identification_example} presents the prompt used along with an illustrative example.

\begin{table*}
    \centering
    \small
    \adjustbox{max width=\textwidth}{
        \begin{tabular}{p{0.3cm}|p{14.7cm}}
            & Prompt used for hallucination identification \\
            \toprule
            \parbox[t]{2mm}{\multirow{3}{*}{1}} & 
            \textbf{System prompt:} You are given a summary and its corresponding source document.  
Your task is to identify the first hallucinated unit -- a word, phrase, or clause that is not clearly supported or is contradicted by the document.
Mark the hallucinated unit in the summary using the following two tags: \\
& - Insert \texttt{[HALLUCINATION\_STARTING\_TAG]} immediately before the first character of the hallucinated unit. \\
& - Insert \texttt{[HALLUCINATION\_TURNOUT\_TAG]} immediately after the last character of the hallucinated unit. \\
& Return the modified summary only, with the tags inserted in the correct positions. \\
& Do not return anything else — no explanations. \\
& If the summary is entirely faithful, return it unchanged, with no tags.\\

 & \textbf{Document:} ... Cardinal Jorge Mario Bergoglio was elected the 256th Pope, President Cristina Kirchner appeared to be gearing up to use Francis I’s powerful new status to Argentina’s advantage ... \\
 & \textbf{Summary:} Pope Francis has been sworn in as the new \textcolor{red}{Pope of Argentina}, a move feared to be a catalyst for nationalism. \\
 & \textbf{GPT-4 output:} Pope Francis has been sworn in as the new \texttt{[HALLUCINATION\_STARTING\_TAG]}Pope of Argentina\texttt{[HALLUCINATION\_TURNOUT\_TAG]}, a move feared to be a catalyst for nationalism. \\
            \bottomrule
        \end{tabular}
    }
    \caption{Prompt for hallucination identification with example.}
    \label{tab:hallucination_identification_example}
\end{table*}

\begin{table*}
    \centering
    \small
    \adjustbox{max width=\textwidth}{
        \begin{tabular}{p{0.3cm}|p{14.7cm}}
            & Prompts used for hallucination generation \\
            \toprule
            \parbox[t]{2mm}{\multirow{2}{*}{1}} & 
            \textbf{System prompt:} You are given a document containing factual information. Your task is to write a concise summary (2–4 sentences) that includes a subtle and believable hallucination. The hallucination should infer something that was not stated in the document but sounds plausible and is difficult to detect. \\
            & Surround only the hallucinated **phrase** - not the entire sentence - with the following tags: \texttt{[HALLUCINATION\_STARTING\_TAG]} ... \texttt{[HALLUCINATION\_TURNOUT\_TAG]}. Return only the modified summary with the hallucination tags inserted. Do not include any additional output or explanation. \\
            
            \midrule
            \parbox[t]{2mm}{\multirow{2}{*}{2}} & 
            \textbf{System prompt:} You are given a document containing factual information. Your task is to write a concise summary (2–4 sentences) that includes a subtle and believable hallucination. The hallucination should misrepresent a minor detail (like date or count). \\
            & Surround only the hallucinated **phrase** — not the entire sentence — with the following tags: \texttt{[HALLUCINATION\_STARTING\_TAG]} ... \texttt{[HALLUCINATION\_TURNOUT\_TAG]}. Return only the modified summary with the hallucination tags inserted. Do not include any additional output or explanation. \\
            
            \midrule
            \parbox[t]{2mm}{\multirow{2}{*}{3}} & 
            \textbf{System prompt:} You are given a document containing factual information. Your task is to write a concise summary (2–4 sentences) that includes a subtle and believable hallucination. This hallucination should introduce an emotional or symbolic interpretation that is not directly supported by the document, yet remains believable within its context. \\
            & Surround only the hallucinated **phrase** — not the entire sentence — with the following tags: \texttt{[HALLUCINATION\_STARTING\_TAG]} ... \texttt{[HALLUCINATION\_TURNOUT\_TAG]}. Return only the modified summary with the hallucination tags inserted. Do not include any additional output or explanation. \\
            
            \midrule
            \parbox[t]{2mm}{\multirow{2}{*}{4}} & 
            \textbf{System prompt:} You are given a document containing factual information. Your task is to write a concise summary (2–4 sentences) that includes a subtle and believable hallucination. The hallucination should combine multiple facts into a single, inaccurate generalization. \\
            & Surround only the hallucinated **phrase** — not the entire sentence — with the following tags: \texttt{[HALLUCINATION\_STARTING\_TAG]} ... \texttt{[HALLUCINATION\_TURNOUT\_TAG]}. Return only the modified summary with the hallucination tags inserted. Do not include any additional output or explanation. \\
            \bottomrule
        \end{tabular}
    }
    \caption{Prompts used for hallucination generation.}
    \label{tab:hallucination_generation_prompts}
\end{table*}

\paragraph{Error analysis of subtle hallucinations}
We fine-tuned a version of \prefixesnlimodelname{} on the hallucination identification data described above. To gain insight into its limitations, we manually analyzed a subset of its predictions on the \ragtruthprefixes{} development set. This qualitative analysis revealed recurring failure cases where the model struggled to detect subtle hallucinations. These included claims that were inferred but not explicitly supported, slight modifications of factual details, and general statements that lacked sufficient grounding in the source. Table~\ref{tab:manual_analysis_examples} presents a representative example of such a challenging case.

\begin{table*}[t]
    \centering
    \small
    \begin{tabular}{lcc}
        \toprule
        \textbf{Method} & \textbf{\summeditsprefixes{}} & \textbf{\ragtruthprefixes{}} \\
        \midrule
        Identification Only & 77.5 $\pm$ 0.8 & 44.7 $\pm$ 0.1 \\
        Identification + Training with Injected Hallucinations (\prefixesnlimodelname{}) & \textbf{78.1} $\pm$ 0.8 & \textbf{47.6} $\pm$ 0.1 \\
        \bottomrule
    \end{tabular}
    \caption{F1 scores for the \textbf{unfaithful} class on the \summeditsprefixes{} and \ragtruthprefixes{} benchmarks. The second model was trained with additional examples containing synthetically injected hallucinations.}
    \label{tab:results_before_injection}
\end{table*}

\begin{table*}
    \centering
    \small
    \adjustbox{max width=\textwidth}{
        \begin{tabular}{p{0.3cm}|p{14.7cm}}
            & Examples from error analysis \\
            \toprule
            \multirow{3}{*}{1} & 
            \textbf{Document:} The FBI charged a Philadelphia woman on Thursday with trying to travel overseas to fight for ISIS. \textcolor{blue}{She’s one of three women arrested this week on terror charges. Two New York women were also taken into custody.} An FBI complaint cites numerous social media messages dating back to August 2013 that were sent by Keonna Thomas, 30, also known as "Young Lioness" and "Fatayat Al Khilafah." One Twitter message said, "If we truly knew the realities ... we all would be rushing to join our brothers in the front lines pray ALLAH accept us as shuhada [martyrs]." Another said, "When you're a mujahid [violent jihadi fighter] your death becomes a wedding." The FBI said Thomas purchased an electronic visa to Turkey on March 23. Turkey is known as the easiest place from which to enter Syria and join ISIS. An ISIS manual advises recruits to buy round-trip tickets to vacation spots such as Spain and then purchase tickets for their real destination once they arrive overseas, the FBI said. On March 26, Thomas purchased a ticket to Barcelona, with a March 29 departure and an April 15 return to the United States, the complaint said. It’s not clear when or where she was arrested. She was charged with knowingly attempting to provide material support and resources to a designated foreign terrorist organization. She could be sentenced to 15 years in prison. \textcolor{blue}{On Thursday, Noelle Velentzas, 28, and her former roommate, Asia Siddiqui, 31, were arrested in New York and accused of planning to build an explosive device for attacks in the United States, federal prosecutors said.} In the past 18 months, the Justice Department’s National Security Division has prosecuted or is prosecuting more than 30 cases of people attempting to travel abroad to join or provide support to terrorist groups. Of those cases, 18 allegedly involve support to ISIS. "The terrorist threat is more decentralized, more diffuse, more complicated," Homeland Security Secretary Jeh Johnson told reporters Thursday. "It involves the potential lone wolf actor, it involves the effective use of social media, the Internet." \\
            & \textbf{Hypothesis:} \textcolor{red}{Three women, including Keonna Thomas of Philadelphia, were charged with attempting to join ISIS.} \\
            & \textbf{Error Analysis:} The hallucination is that three women were charged with attempting to join ISIS, but only Keonna Thomas was charged with that; the other two were arrested on separate terror-related charges. \\
            \bottomrule
        \end{tabular}
    }
    \caption{Examples of errors caused by the identification-only version of \prefixesnlimodelname{}.}
    \label{tab:manual_analysis_examples}
\end{table*}

Motivated by the findings from our error analysis, we constructed a synthetic dataset containing subtle hallucinations that the model previously failed to detect. These examples were generated through targeted prompting of OpenAI’s GPT-4 (O3) model \citep{openai2024gpt4technicalreport}; see Table~\ref{tab:hallucination_generation_prompts} for the prompts and Table~\ref{tab:generation_example} for an illustrative example. This dataset introduced a wider variety of fine-grained hallucination types into training. As shown in Table~\ref{tab:results_before_injection}, incorporating this data led to improved model performance.
While the identification dataset contains some naturally occurring non-consecutive hallucinations, our prompts primarily target consecutive ones.

\begin{table*}
    \centering
    \small
    \adjustbox{max width=\textwidth}{
    \begin{tabular}{p{0.3cm}|p{14.7cm}}
        & Hallucination generation example \\
        \toprule
        \parbox[t]{2mm}{\multirow{5}{*}{1}} & 
        \textbf{System prompt:} You are given a document containing factual information. Your task is to write a concise summary (2–4 sentences) that includes a subtle and believable hallucination. The hallucination should misrepresent a minor detail (like date or count). \\
            & Surround only the hallucinated **phrase** — not the entire sentence — with the following tags: \texttt{[HALLUCINATION\_STARTING\_TAG]} ... \texttt{[HALLUCINATION\_TURNOUT\_TAG]}. Return only the modified summary with the hallucination tags inserted. Do not include any additional output or explanation. \\

        & \textbf{Document:} ... both parties could expect their share of the national vote to count towards a representative share of seats in the \textcolor{blue}{120-member} parliament ... \\
        
        & \textbf{GPT-4 output:} The unlikely alliance between Dotcom’s Internet Party and Maori nationalist-led Mana Party seeks to challenge New Zealand’s electoral system by combining contrasting approaches to internet freedom and indigenous representation. Political scientist Bryce Edwards noted that this partnership could secure success in the September elections, especially if their vote share translates into seats in a \texttt{[HALLUCINATION\_STARTING\_TAG]}\textcolor{red}{170-member} parliament\texttt{[HALLUCINATION\_TURNOUT\_TAG]}. \\

        \bottomrule
    \end{tabular}
    }
    \caption{Example of hallucination generation by misrepresenting a minor numerical detail.}
    \label{tab:generation_example}
\end{table*}

The \trueteacher{} dataset \citep{gekhman2023trueteacherlearningfactualconsistency} is released under the CC-BY-NC 4.0 license. As our prefix-level entailment training dataset is derived from it, we will also release it under the same CC-BY-NC 4.0 license.

\section{Evaluation Benchmark Datasets}\label{appendix:eval_datasets}
Table~\ref{tab:more_data_stats} provides statistics for the evaluation benchmark datasets.
The \ragtruthprefixes{} dataset is released under the same MIT license as the original \ragtruth{} dataset \cite{niu2024ragtruthhallucinationcorpusdeveloping}. Similarly, the \summeditsprefixes{} dataset follows the CC-BY 4.0 license of the original \summedits{} dataset \cite{laban2023llmsfactualreasonersinsights}.
\section{Model Training Details}
\label{appendix:more_train_details}
We provide additional training details for our \prefixesnlimodelname{} model, supplementing the information in \S\ref{sec:nli_model}.
We used the following prompt, following \trueteacher{} \citep{gekhman2023trueteacherlearningfactualconsistency} for the entailment prediction:
\textit{\textbf{Premise}: \{document\} \textbf{Hypothesis}: \{summary prefix\}}.
The model is trained to predict ``1'' if the hypothesis is factually consistent and ``0'' otherwise. We infer that the hypothesis is entailed if the probability for the token ``1'' is higher than 0.5, otherwise we deduce that it is not entailed. 

For fine-tuning \minitrue{}, we used a learning rate of $2 \times 10^{-4}$ and a batch size of 32, training for 3 epochs with LoRA. For \prefixesnlimodelname{}, we used a lower learning rate of $5 \times 10^{-6}$, the same batch size of 32, and also trained for 3 epochs using LoRA. During training, we set a maximum input length of 2048 tokens. In line with our goal of detecting inconsistencies, we selected the checkpoint with the highest F1 score on the unfaithful class over our development set.

To preserve \prefixesnlimodelname{} pre-trained entailment knowledge while adapting it to the new setting of truncated hypotheses, we froze all layers except the final one. We also experimented with full fine-tuning but found that freezing all but the last layer yielded better results. This approach allowed the model to retain its foundational understanding of entailment while improving its capacity to assess incomplete text.

\section{More Results for \prefixesnlimodelname{}} \label{appendix:results}
\subsection{\prefixesnlimodelname{} } \label{appendix: mtp_additional_metrics}

\begin{table}[t]
    \centering
    \small
    \adjustbox{max width=\columnwidth}{

    \begin{tabular}{lcc}
        \toprule
        \textbf{Model} & \textbf{\summeditsprefixes{}} & \textbf{\ragtruthprefixes{}} \\
        \midrule
        \minicheck{} (Flan-T5) & 76.8 $\pm$ 0.9 & 89.7 $\pm$ 0.1 \\
        \minitrue{} & 78.7 $\pm$ 0.8 & 91.0 $\pm$ 0.1 \\
        \prefixesnlimodelname{} & \textbf{81.4} $\pm$ 0.8 & \textbf{95.0} $\pm$ 0.1 \\
        \bottomrule
    \end{tabular}}
    \caption{F1 scores for the faithful class on the \summedits{} and \ragtruth{} prefixes benchmarks.}
    \label{tab:mtp_faithful_f1}
\end{table}

We previously reported F1 for the unfaithful class (Section~\ref{sec:prefix_results}), reflecting our emphasis on detecting hallucinations. In this section, we provide complementary results for the faithful class, for control. Specifically, we report micro-averaged F1 scores in Table~\ref{tab:mtp_faithful_f1} as a control metric.

As shown in the table, \prefixesnlimodelname{} consistently outperforms both baselines across the two benchmarks, with a 2.7-point gain over its base model \minitrue{} on \summeditsprefixes{} and a 4-point improvement on \ragtruthprefixes{}.

\subsection{\minitrue{} performance over the TRUE benchmark} \label{appendix:minitrue_true}

To develop a lightweight entailment model, we trained \minitrue{} on the same dataset as \trueteacher{} \citep{gekhman2023trueteacherlearningfactualconsistency} (see Section~\ref{subsec:training_regime} for more training details). We assessed its ability to detect factual inconsistencies by evaluating it on the summarization subset of the TRUE benchmark \citep{honovich2022truereevaluatingfactualconsistency}.

As shown in Table~\ref{tab:minitrue_truebenchmark}, \minitrue{} achieves performance that is comparable to \trueteacher{}, despite being significantly smaller in size. These results support the suitability of \minitrue{} as a lightweight alternative for entailment models.

\begin{table*}[!htbp]
\begin{center}
\resizebox{\textwidth}{!}{
\begin{tabular}{l|rrrrr|r}
\toprule
\multicolumn{1}{c|}{} &
  \multicolumn{1}{c}{\bf MNBM} &
  \multicolumn{1}{c}{\bf QAGS-X} &
  \multicolumn{1}{c}{\bf FRANK} &
  \multicolumn{1}{c}{\bf SummEval} &
  \multicolumn{1}{c}{\bf QAGS-C} &
  \multicolumn{1}{|c}{\bf Average} \\
\midrule
\trueteacher{} (11B) \citep{gekhman2023trueteacherlearningfactualconsistency}*                        & 78.1 & \bf 89.4 & \bf 93.6 & \bf 88.5 & \bf 89.4 & \bf 87.8 \\
\minitrue{} (1B) & \textbf{79.2}& 85.1 & 90.3& 83.2 & 88.6 & 85.3 \\
\bottomrule
\end{tabular}
}
\end{center}

\caption{ROC-AUC results on the summarization subset of the TRUE benchmark \cite{honovich2022truereevaluatingfactualconsistency}. Results for \trueteacher{} were reported in \citet{gekhman2023trueteacherlearningfactualconsistency}.}
\label{tab:minitrue_truebenchmark}
\end{table*}
\subsection{Successful detection example}\label{appendix:qualitative_examples}

Table~\ref{tab:mtp_prefix_success} presents a case where \prefixesnlimodelname{} successfully detected a subtle hallucination, while both \minitrue{} and \minicheck{} failed. This may be because these models were trained exclusively on full sentences and are less effective when the input is not a complete sentence. In contrast, \prefixesnlimodelname{}, trained on prefix-level entailment inputs, is better suited to handle incomplete or partial inputs, enabling it to detect hallucinations in prefixes more reliably.
\begin{table}[!htbp]
\centering
\small
\begin{tabular}{p{0.93\linewidth}}
\toprule
\textbf{Premise (truncated)} \\
\midrule
... John T. Booker Jr. of Topeka, an American citizen also known as Mohammed Abdullah Hassan, was taken into custody near Manhattan, Kansas, in a van that contained what he thought was a bomb ... \textcolor{blue}{Booker enlisted in the Army last year} and was due to ship out to basic training \textcolor{blue}{April 7, 2014} ... \textcolor{blue}{His enlistment was terminated March 24, 2014}, at the request of Army Criminal Investigation Command ... \\
\midrule
\textbf{Hypothesis} \\
\midrule
A \textcolor{red}{US Army veteran} has been arrested \\
\midrule
\textbf{Gold Label} \\
\midrule
\textsc{Not Entailed} \\
\midrule
\textbf{Rationale} \\
\midrule
The hypothesis refers to Booker as a "US Army veteran," which implies he completed service. However, the premise states his enlistment was terminated before training began. He never served and thus is not a veteran. \\
\bottomrule
\end{tabular}
\caption{An example where \prefixesnlimodelname{} correctly detects a factual hallucination describing Booker as a “US Army veteran” that \minitrue{} and \minicheck{}, which were trained on full sentences, fail to catch.}
\label{tab:mtp_prefix_success}
\end{table}

\subsection{Error analysis for \prefixesnlimodelname{}}\label{appendix:error_analysis}
To better understand the failure modes of \prefixesnlimodelname{} on \ragtruthprefixes{}, we conducted a manual analysis of 60 misclassified examples: 30 false positives (FPs) and 30 false negatives (FNs).

Each instance was assigned to a single error category based on the primary cause of the model's mistake. Tables~\ref{tab:fp_distribution} and~\ref{tab:fn_distribution} report the distribution and explanations for each category.

\paragraph{Prefix incompleteness is not a dominant failure mode.}
Only 3.3\% of FPs and 13.3\% of FNs were caused by underdetermined prefixes, cases where the input lacked sufficient context to determine entailment. This suggests that prefix incompleteness is not the main cause of model errors, and that in most cases, the prefix alone provides enough information for a correct decision.

\paragraph{Generic misclassification is the most frequent source of error.}
The largest share of FPs (53.3\%) and a substantial portion of FNs (20\%) fall into the “Generic Prediction Error” category, referring to cases where the input was clear but the model’s prediction was plainly incorrect. These types of errors are also common in standard NLI models, indicating that they reflect general prediction challenges rather than prefix-specific issues.

\paragraph{Shallow heuristics drive many false positives.}
A significant portion of false positives (33.3\%) arise from surface-level heuristic biases. These errors may indicate limited semantic reasoning, where shallow lexical or structural patterns are favored over actual consistency. Mitigating this bias could lead to more reliable models.

\paragraph{False negatives reveal reasoning gaps.}
The most common false negative category (43.3\%) involves missed inferences, where the entailment may be present but not explicitly stated, potentially requiring temporal, causal, or commonsense reasoning. This pattern suggests that the model may struggle to capture implicit meaning or integrate less directly stated information.

\paragraph{Sensitivity to paraphrase and implicit knowledge remains limited.}
13.3\% of false negatives involved surface-level mismatches, where paraphrastic entailment was not recognized, and an additional 3.3\% involved entailments that may rely on world knowledge. These cases point to potential challenges in generalizing beyond literal overlap and handling implicit meaning.

These findings suggest that many of the errors made by \prefixesnlimodelname{} reflect general challenges commonly seen in NLI models, including difficulties with semantic reasoning, inference, and robustness to paraphrasing or lexical variation, rather than issues specific to using prefix hypotheses as input.

\begin{table*}[t]
\centering
\small
\begin{tabular}{c p{3.5cm} p{9.5cm} c}
\hline
\textbf{\#} & \textbf{Category} & \textbf{Explanation} & \textbf{FP (\%)} \\
\hline
1 & Underdetermined Prefix & The prefix is too short, ambiguous, or lacks necessary context. False positives may result from the limited information available in the prefix. & 3.33\% \\
\hline
2 & Surface-Level Heuristic Bias & The prefix shares significant surface-level lexical or structural features with the source, which may lead the model to predict entailment despite the absence of true semantic consistency. & 33.33\% \\
\hline
3 & World Knowledge Confusion & Cases where entailment may depend on somewhat specific world knowledge, which was not assumed in the gold annotation and therefore labeled as not entailed. & 3.33\% \\
\hline
4 & Generic Prediction Error & The prefix is understandable, the source is clear, but the model gets it wrong. & 53.33\% \\
\hline
5 & Wrong Annotation & The model prediction is reasonable, but the gold annotation is incorrect. & 6.67\% \\
\hline
\end{tabular}
\caption{Distribution of false positives (FP) by error category.}
\label{tab:fp_distribution}
\end{table*}

\vspace{1em}

\begin{table*}[t]
\centering
\small
\begin{tabular}{c p{3.5cm} p{9.5cm} c}
\hline
\textbf{\#} & \textbf{Category} & \textbf{Explanation} & \textbf{FN (\%)} \\
\hline
1 & Underdetermined Prefix & The prefix is too short, ambiguous, or lacks necessary context. False negatives occur when the model hesitates due to missing context despite underlying entailment. & 13.33\%\\
\hline
2 & Incorrect Reasoning / Inference & The hypothesis is clear and understandable, but the prefix contains information that may call for implicit reasoning (e.g., temporal, causal, or logical). The model does not predict entailment, possibly due to the complexity of the inference. & 43.33\% \\
\hline
3 & Surface-Level Heuristic Bias & The hypothesis is entailed but paraphrased with limited surface similarity to the premise, which may challenge the model when predictions depend on more than lexical or structural overlap. & 13.33\% \\
\hline
4 & World Knowledge Confusion & Cases where the gold annotator assumes access to somewhat specific world knowledge, but the model does not predict entailment. The missed entailment may reflect challenges in applying or accessing such knowledge. & 3.33\% \\
\hline
5 & Generic Prediction Error & The prefix is understandable, the source is clear, but the model gets it wrong. & 20.0\% \\
\hline
6 & Wrong Annotation & The model prediction is reasonable, but the gold annotation is incorrect. & 6.67\% \\
\hline
\end{tabular}
\caption{Distribution of false negatives (FN) by error category.}
\label{tab:fn_distribution}
\end{table*}

\section{Controlled Decoding Experimental Setup}
\label{appendix:additional_cd_experimental_setup}

\subsection{Penalty design} \label{appendix:rectification_threshold}

\begin{table}[t] 
\centering
\begin{adjustbox}{width=\columnwidth} 
\begin{tabular}{lrrrr} 
\toprule
 & \multicolumn{2}{c}{CNN/DailyMail} & \multicolumn{2}{c}{XSUM} \\
\cmidrule(lr){2-3} \cmidrule(lr){4-5} 
Threshold & MiniCheck & ROUGE-L & MiniCheck & ROUGE-L \\
\midrule
0.1       &  78.19 &   16.97 &  72.76 &   13.59 \\
0.2       &  78.71 &   16.86 &  74.32 &   13.68 \\
0.3       &  79.35 &   16.85 &  74.58 &   13.55 \\
0.4       &  80.33 &   16.80 &  76.13 &   13.72 \\
\textbf{0.5} & \textbf{80.72} & \textbf{16.93} & \textbf{76.30} & \textbf{13.58} \\
0.6       &  77.09 &   16.80 &  74.49 &   13.65 \\
0.7       &  69.86 &   15.78 &  60.23 &   12.99 \\
0.8       &  64.52 &   15.50 &  58.03 &   12.72 \\
0.9       &  66.73 &   15.02 &  55.33 &   12.11 \\
1.0       &  87.90 &   15.99 &  79.40 &   12.08 \\
\bottomrule
\end{tabular}
\end{adjustbox}
\caption{
Tuning results for CNN/DM and XSUM.
}
\label{tab:threshold-tuning} 
\end{table}

We tuned the rectification threshold ($\tau$) on 500 development samples from both CNN/DM and XSUM, with results shown in Table~\ref{tab:threshold-tuning}. Our objective was to maximize MiniCheck (our faithfulness metric) without degrading the ROUGE-L control metric. We selected $\tau=0.5$, as it achieved the highest MiniCheck scores (80.72 CNN/DM, 76.30 XSUM) within the range of stable ROUGE-L performance. This value was used for all subsequent experiments.

The underlying rationale of our penalty design was to integrate the PrefixNLI model’s probability into the decoder’s logits on a compatible numerical scale, achieved through a log-odds transformation. This transformation penalizes probabilities below 0.5 while rewarding those above it, unlike log-probabilities, which are always negative. In early experiments without rectification, where the signal also amplified highly faithful prefixes, we found that such amplification interfered with the decoder’s inherent ability to produce high-quality text, resulting in visible degradations in summary quality and reduced ROUGE scores. Consequently, we adopted a rectification strategy that intervenes only for prefixes with low entailment probabilities, thereby discouraging unfaithful continuations rather than amplifying faithful ones. The rectification threshold $\tau$ determines the intervention strength, and its optimal value (0.5) was empirically identified as balancing faithfulness improvements with maintained summary quality.

\subsection{Hyper-parameters used}\label{appendix:cd_hyperparameters}

Results are reported for a beam size $K=3$. For generation, we use $p=0.9$ for considering the top-$p$ candidates in the beam (the most likely tokens whose accumulated probability reaches $p$). For computational efficiency, based on tuning performed on the development set, we limit this candidate set to a maximum number of tokens per beam. This limit is set to $20$ for the LLaMA models, $10$ for OLMo-1B, and $7$ for OLMo-7B.

We conducted hyperparameter tuning using 500 instances from the CNN/DM development set, evaluating values from 1 to 15 in increments of 2 as the scaling factor. Our experiments identified $\lambda = 5$ as optimal, producing the most faithful results.

\subsection{Summaries generation prompt} \label{appendix:generation_prompt_cd}
For our controlled decoding experiment, described in Section~\ref{subsec:cd_experimental_setup}, we used the following prompt for the generation of summaries:
\textit{Summarize the following text accurately and concisely. Output only the summary---do not include any introductory words like 'Summary:' or explanations.}

\subsection{CAD experimental setup}\label{appendix:cad_details}
In Section~\S\ref{sec:cd}, we use Context-Aware Decoding (CAD)~\citep{shi-etal-2024-trusting} as a baseline. We adopt the hyperparameters recommended by the authors, setting $\alpha = 0.5$ and using top-$p$ sampling with $p = 0.9$ for summarization. The experiment is conducted with the \texttt{LLaMA-3.2-1B-Instruct} model, using the document as context, consistent with the original CAD setup. The input format was:

\begin{quote}
\texttt{\{"role": "system", "content": "Summarize the following text accurately and concisely. Output only the summary—do not include any introductory words like 'Summary:' or explanations."\}},\\
\texttt{\{"role": "user", "content": user\_content\}}
\end{quote}

For the no-context condition, we set \texttt{user\_content} to \texttt{[Text omitted]}. For the context condition, we provided the full document.

\subsection{GPT-4 faithfulness metric}\label{appendix:gpt_faithfulness}
The prompt used for evaluating the faithfulness of the generated summaries from Section~\ref{sec:cd} can be found in Table~\ref{tab:faithfulness_prompt}.
We used gpt-4.1-2025-04-14.
\begin{table*}
    \centering
    \small
    \adjustbox{max width=\textwidth}{
        \begin{tabular}{p{15cm}}
            \textbf{System prompt:} Determine whether the provided summary is consistent
with the corresponding document. Consistency in this
context implies that all information presented in the
response is substantiated by the document. If not, it should
be considered inconsistent.

The response can have one or more of the following errors:  
1. Extrinsic Information: the response contains new
information not grounded in the source material  
2. Mis-Referencing: a property or an event in the response
can be found in the source material, but are associated
with the wrong entity  
3. Stating Opinion As Fact: the response entails a
proposition that’s mentioned in the source material not as
a fact, but as someone’s opinion  
4. Reasoning Error: the response makes one or more
wrong inferences from the information in the source
material  
5. Tense/modality Error: the tense or modal (e.g., can, may,
must) used in the response sentence does not match the
tense/modality of the source material  
6. Contradiction: the response contradicts the source
material  
7. Nuanced Meaning Shift: the response twists information
from the source material in a subtle way  

Given the error categories, rate the above response on a
scale of 1 to 5 based on extent of factual consistency:  
5. completely consistent: the response is completely factually consistent with the source material.  
4. insignificant inconsistencies: the response is mostly
factually consistent, with slight inconsistencies not
affecting main points.  
3. partially inconsistent: overall factually consistent, with
a few inconsistencies with the source material.  
2. severe inconsistencies: nearly half response is factually
inconsistent, with severe deviation from main points.  
1. completely inconsistent: the entire response is factually
inconsistent with the source material.  

First output a list of errors that the summary makes, then
conclude the response with a score in the following format:  
"therefore, the score is:"
            \\
        \end{tabular}
    }
    \caption{Prompt \citep{wadhwa-etal-2024-learning-refine} used for evaluating faithfulness of the generated summaries from Section~\ref{sec:cd}.}
    \label{tab:faithfulness_prompt}
\end{table*}

\subsection{Additional details on the experimental setup}\label{appendix:exp_setup_cd}
Our controlled decoding method (Section~\ref{subsec:cd_method}) frequently invokes the entailment model to score summary prefixes. For efficient inference, we run the entailment model using vLLM \citep{kwon2023efficient}.
Summary generation across all experiments required approximately 45 GPU hours on an NVIDIA A100-80GB.

\section{Ablation Study}\label{appendix:ablation_study}
We report results when applying the \minitrue{} model to beam prefixes, allowing us to assess the relative benefits of \prefixesnlimodelname{} over \minitrue{}, which was trained only on complete-sentence hypotheses (\S\ref{sec:nli_model}). As shown in Table~\ref{tab:ablation_results}, \prefixesnlimodelname{} consistently outperforms \minitrue{} across all evaluation metrics and model sizes, with more than a 2-point improvement in the \minicheck{} faithfulness score across both datasets and all model scales. Notably, applying \minitrue{} with the 8B model on XSum degraded performance relative to the vanilla model, which achieved a \minicheck{} score of 81.5 (see Table~\ref{tab:cd_results}). These findings highlight the necessity of a dedicated prefix-level entailment model, rather than relying on NLI models trained exclusively on complete-sentence hypotheses.

\begin{table*}[!htbp]
    \centering
    \small
    \resizebox{\textwidth}{!}{
    \begin{tabular}{l | cc | ccc || cc | ccc}
        \toprule
        \multirow{3}{*}{Model} 
        & \multicolumn{5}{c||}{\textbf{XSum}} 
        & \multicolumn{5}{c}{\textbf{CNN/DM}} \\
        \cmidrule{2-11}
        & \multicolumn{2}{c|}{Faithfulness} & \multicolumn{3}{c||}{Auxiliary Metrics} 
        & \multicolumn{2}{c|}{Faithfulness} & \multicolumn{3}{c}{Auxiliary Metrics} \\
        & \minicheck{}$\uparrow$ & GPT-4$\uparrow$ & R-L$\uparrow$ & \mauve{}$\uparrow$ & Speed$\downarrow$
        & \minicheck{}$\uparrow$ & GPT-4$\uparrow$ & R-L$\uparrow$ & \mauve{}$\uparrow$ & Speed$\downarrow$ \\
        \midrule
        Prefix (1B, MT) 
                    & 72.4 $\pm$ 0.6 & 3.09 $\pm$ 0.03 & 13.5 $\pm$ 0.1 & 89.6 & 4.89 
                    & 77.2 $\pm$ 0.6 & 3.36 $\pm$ 0.02 & 19.4 $\pm$ 0.1 & \textbf{79.2} & 5.69 \\
        Prefix(3B, MT)
                    & 79.2 $\pm$ 0.6 & 3.54 $\pm$ 0.03 & 14.2 $\pm$ 0.1 & 91.1 & 5.27
                    & 82.3 $\pm$ 0.7 & 3.71 $\pm$ 0.02 & \textbf{19.8} $\pm$ 0.1 & 79.1 & 6.62 \\
        Prefix(8B, MT)
                   & 79.7 $\pm$ 0.6 & 3.75 $\pm$ 0.03 & \textbf{14.6} $\pm$ 0.1 & 91.2 & 7.18
                   & 88.7 $\pm$ 0.4 & 3.85 $\pm$ 0.02 & 18.9 $\pm$ 0.1 & 79.1 & 9.35\\
        \midrule
        Prefix (1B, \shortprefixesnlimodelname{}) 
            & 74.8 $\pm$ 0.6 & 3.13 $\pm$ 0.03 & 13.6 $\pm$ 0.1 & 89.6 & \textbf{4.84} 
            & 79.2 $\pm$ 0.6 & 3.41 $\pm$ 0.02 & \textbf{19.8} $\pm$ 0.1 & 79.1 & \textbf{5.67} \\
        Prefix (3B, \shortprefixesnlimodelname{}) 
            & 82.4 $\pm$ 0.5 & 3.59 $\pm$ 0.03 & 14.1 $\pm$ 0.1 & 91.2 & 5.43 
            & 85.8 $\pm$ 0.4 & 3.73 $\pm$ 0.02 & 19.0 $\pm$ 0.1 & 79.1 & 7.18 \\
        Prefix (8B, \shortprefixesnlimodelname{}) 
            & \textbf{87.0} $\pm$ 0.4 & \textbf{3.84} $\pm$ 0.03 & 14.5 $\pm$ 0.1 & \textbf{91.3} & 7.23 
            & \textbf{90.8} $\pm$ 0.3 & \textbf{3.90} $\pm$ 0.02 & 18.9 $\pm$ 0.1 & 79.1 & 9.20 \\

        \bottomrule
    \end{tabular}
    }
    \caption{
        Performance across XSum and CNN/DM datasets. Prefix is used to denote our proposed controlled decoding method, and we abbreviate \textbf{\prefixesnlimodelname{}} as \textbf{\shortprefixesnlimodelname{}} and \textbf{\minitrue{}} as \textbf{MT}. 
        }
    \label{tab:ablation_results}
\end{table*}

\section{Computational Cost Analysis} \label{appendix:comp_cost_analysis}

We estimate the theoretical increase in floating-point operations (FLOPs) incurred by incorporating \textit{PrefixNLI} control into the decoding process, relative to vanilla autoregressive generation. Following the transformer compute formulation of \citet{kaplan2020scalinglawsneurallanguage}, the per-token forward-pass cost can be expressed as:

\[
C_{\text{forward}} \approx 2N + 2n_{\text{layer}} n_{\text{ctx}} d_{\text{model}},
\]
where \(N\) denotes the number of non-embedding parameters, \(n_{\text{layer}}\) is the number of layers, \(n_{\text{ctx}}\) the current context length, and \(d_{\text{model}}\) the hidden dimension. When \(d_{\text{model}} \gg n_{\text{ctx}} / 12\), the attention-dependent term becomes relatively small, and the total per-token cost can be well-approximated by \(C_{\text{forward}} \approx 2N\) \citep{kaplan2020scalinglawsneurallanguage}.

Let \(N_{\text{LM}}\) be the number of non-embedding parameters of the generator language model, and \(N_{\text{ent}}\) that of the entailment model used for \textit{PrefixNLI} control (\(\approx1.23\) B parameters for \prefixesnlimodelname{}, which is a finetuned LLaMA-3.2-1B-Instruct). During decoding, the entailment model evaluates \(M\) alternative next-token candidates for each beam. The additional theoretical compute cost per decoding step per beam is thus:

\[
\text{FLOPs}_{\text{ent}} \approx 2 N_{\text{ent}} M.
\]

Hence, the overall forward-pass compute for one generated token under \textit{PrefixNLI} control becomes:

\[
\text{FLOPs}_{\text{PrefixNLI}} \approx 2N_{\text{LM}} + 2N_{\text{ent}} M.
\]

In our experiments with the LLaMA-3.2-1B-Instruct generator, each beam evaluated up to 20 candidate tokens per step, although the empirically observed average number of candidates under top-\(p\) sampling was about 6.

\paragraph{Prefix caching}
The above estimate already assumes prefix key--value (KV) caching, such that the entailment model does not re-encode the entire prefix for each candidate. At every decoding step, all \(M\) candidate continuations within a beam share the same prefix (the source document and tokens generated so far) and differ only in their proposed next token. The model computes and caches the hidden states for this shared prefix once, and performs \(M\) incremental forward passes, one for each candidate token, on top of the cached representations. After a token is selected and appended to the sequence, the KV cache is extended once and reused for the next decoding step. The prefill step that encodes the source document and initial prefix is executed once at the beginning of generation and remains in GPU memory throughout decoding.

\paragraph{Runtime considerations}
While FLOPs provide a hardware-agnostic estimate of computational cost, they do not fully capture runtime efficiency. In our setup, \textit{\prefixesnlimodelname{}} entailment evaluations are executed through vLLM, which performs optimized batched inference and reuses prefix computations to avoid redundant processing. As a result, the observed wall-clock overhead is substantially lower than what the theoretical FLOPs ratio alone would suggest. Therefore, both FLOPs estimates and empirical runtime measurements (reported in the main text) should be considered when evaluating computational efficiency.

\paragraph{Theoretical vs. practical cost}
Both the generator and entailment models are based on the LLaMA-3.2-1B-Instruct architecture, each containing approximately \(1.23\times10^{9}\) non-embedding parameters. 
Under the \(2N\) approximation, the per-token cost of vanilla decoding is thus
\(\text{FLOPs}_{\text{vanilla}} \approx 2N_{\text{LM}} = 2.46\times10^{9}\) FLOPs.
With \textit{PrefixNLI} control, where the entailment model evaluates an average of \(M=6\) candidates per step, the cost becomes
\(\text{FLOPs}_{\text{PrefixNLI}} \approx 2N_{\text{LM}} + 2N_{\text{ent}}M = 17.22\times10^{9}\) FLOPs,
corresponding to roughly a \(7\times\) theoretical increase relative to vanilla decoding.
However, this ratio represents an upper bound that does not reflect the efficiency gains of prefix caching and vLLM’s batched inference.
Empirically, the measured generation latency on a single NVIDIA~A100~80GB GPU was only about \(2.4\times\) slower than vanilla decoding for the same model,
indicating that much of the additional theoretical compute is effectively mitigated by these optimizations in practice.

\section{Example Outputs} 
\subsection{Controlled decoding generation examples}\label{appendix:improvement_examples}
Table~\ref{tab:example_predictions} presents example summaries generated from XSum documents using our controlled decoding method (Section~\ref{sec:cd}), alongside outputs from the vanilla model. As demonstrated here, our method effectively avoids hallucinations, thereby improving faithfulness.
\begin{table*}
    \centering
    \small
    \adjustbox{max width=\textwidth}{
        \begin{tabular}{p{0.3cm}|p{14.7cm}}
            & Examples \\
            \toprule
            \parbox[t]{2mm}{\multirow{3}{*}{1}} & 
            \textbf{Source document:} The 23 year-old stroked the GB boat to a fast start and they were ahead after 500m but European champions, Romania and New Zealand proved too strong. The Romanians came out on top, with the \textcolor{blue}{British eight comfortably holding off the challenge from the Netherlands}.
            The GB eight now prepare for the World Championships in Sarasota in September. There was no medal success for Enniskillen's Holly Nixon in the women's quadruple sculls. Nixon and her GB team mates Bethany Bryan, Alice Baatz and Mathilda Hodgkins-Byrne, were always off the pace and finished fifth in a race won by Poland. \\
 \\
 & \textbf{Vanilla generated summary:} GB's GB boat took an early lead, but Romania, New Zealand, and the Netherlands overpowered them. Romania won, with \textcolor{red}{GB finishing second behind the Netherlands}.
GB will compete in the 2024 World Championships.
Holly Nixon's GB women's quad team finished fifth.
 \\
            
            \\
    & \textbf{Prefix Entailment (\prefixesnlimodelname{}) generated summary:} GB boat crew ahead after first 500 meters, but Romania and NZ proved too powerful.
Romanians win, \textcolor{darkgreen}{British eight hold off Netherlands challenge}.
GB eight prepare for World Championships.
Holly Nixon's GB team finishes fifth in women's quad sculls.
 \\
            
            \\
            & \textbf{Hallucination Explanation:} GB can’t be second if it is behind the Netherlands and Romania won. \\
            \midrule
            \parbox[t]{2mm}{\multirow{3}{*}{2}} & 
            \textbf{Source document:} The 32-year-old had been playing in the Isthmian Premier Division with Leatherhead following his release by Newport at the end of last season. Pidgeley has made 260 appearances in spells with nine clubs, including Chelsea, Watford and Millwall. Forest Green are currently second in the National League table, one point behind leaders Cheltenham Town. Pidgeley could make his Rovers debut when they host Aldershot on Friday. \\
            \\
             & \textbf{Vanilla generated summary:} Leatherhead's Pidgeley is 32, has played 260 league games, and \textcolor{red}{currently plays for Forest Green}. \\
            \\
            & \textbf{Prefix Entailment (\prefixesnlimodelname{}) generated summary:} Leatherhead player Pidgeley is 32 and has played for nine clubs. He has made over 260 league appearances. Forest Green are second in their National League division.
 \\
            \\
            & \textbf{Hallucination Explanation:} The source document doesn’t mention that Pidgeley currently plays for Forest Green. \\
            \bottomrule
        \end{tabular}
    }
    \caption{Faithfulness improvement examples achieved through our PrefixNLI-based controlled decoding method. In the first example, the vanilla summary contradicts the source, while also being self-contradictory, whereis our method's summary states the correct fact. In the second example the vanilla summary hallucinates a fact that is not present in the source (a neutral entailment case), which our method's summary avoids.}
    \label{tab:example_predictions}
\end{table*}

\subsection{XSum's reference summaries are not faithful} \label{appendix:xsum_hallucination_example}

As discussed in Section~\ref{subsec:cd_results}, our method resulted in a lower ROUGE-L score on the XSum dataset. We attribute this decline to the fact that our method encourages faithfulness, while XSum’s reference summaries frequently contain hallucinated content. Table~\ref{tab:example_xsum_hallucination} illustrates an example of such a hallucination.

This issue stems from the design of the XSum dataset: the reference summaries are not actual summaries, and were not required to be faithful. Prior analysis has shown that over 70\% of XSum’s reference summaries contain factual inconsistencies \citep{maynez-etal-2020-faithfulness}.

\begin{table*}
    \centering
    \small
    \adjustbox{max width=\textwidth}{
        \begin{tabular}{p{0.3cm}|p{14.7cm}}
            & Examples \\
            \toprule
            \parbox[t]{2mm}{\multirow{3}{*}{1}} & 
            \textbf{Source document:} It happened on the Linn Road in the town at about 14:05 BST on Sunday.
            The two men have been taken to hospital for treatment for their injuries.
            Police have appealed for anyone with information about the attack to contact them. \\
 \\
 & \textbf{Reference summary:} Two men have been assaulted \textcolor{red}{in Larne, County Antrim, by a gang armed with baseball bats and a hatchet}.
 \\

            \\
            & \textbf{Prefix Entailment (\prefixesnlimodelname{}) generated summary}: Men attacked on Linn road, taken to hospitals for treatment. Police appeal for information
            \\
            \\
            & \textbf{Hallucination Explanation:} The source document doesn't mention that Linn road is located in Larne. Additionally, the source document doesn't mention that the attackers were armed with baseball bats and a hatchet. \\
            \bottomrule
        \end{tabular}
    }
    \caption{Examples showing XSum's reference summaries aren't faithful.}
    \label{tab:example_xsum_hallucination}
\end{table*}
\end{document}